# *Point2Graph*: An End-to-end Point Cloud-based 3D Open-Vocabulary Scene Graph for Robot Navigation

Yifan Xu, Ziming Luo*, Qianwei Wang*, Vineet Kamat, and Carol Menassa

*Abstract*— Current open-vocabulary scene graph generation algorithms highly rely on both 3D scene point cloud data and posed RGB-D images and thus have limited applications in scenarios where RGB-D images or camera poses are not readily available. To solve this problem, we propose *Point2Graph*, a novel end-to-end point cloud-based 3D open-vocabulary scene graph generation framework in which the requirement of posed RGB-D image series is eliminated. This hierarchical framework contains room and object detection/segmentation and open-vocabulary classification. For the room layer, we leverage the advantage of merging the geometry-based border detection algorithm with the learning-based region detection to segment rooms and create a "Snap-Lookup" framework for open-vocabulary room classification. In addition, we create an end-to-end pipeline for the object layer to detect and classify 3D objects based solely on 3D point cloud data. Our evaluation results show that our framework can outperform the current state-of-the-art (SOTA) open-vocabulary object and room segmentation and classification algorithm on widely used real-scene datasets. We provide code and videos at: https://point2graph.github.io/

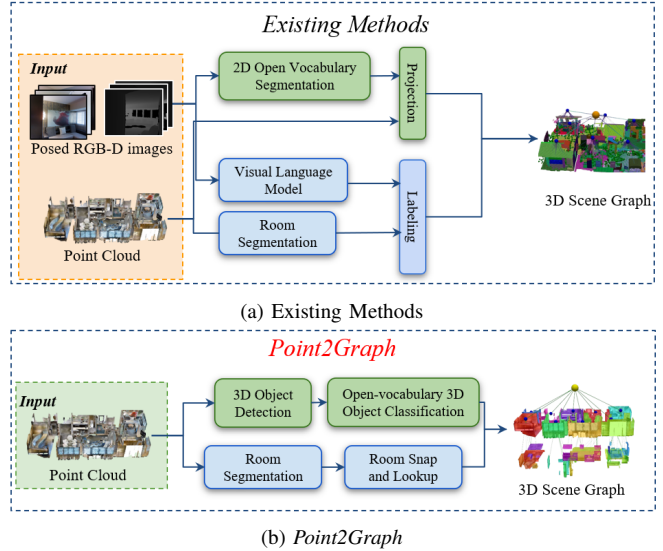

(a) Existing Methods

(b) *Point2Graph*

Fig. 1: **Comparison of our proposed *Point2Graph* algorithm and the existing 3D scene graph generation algorithm.** Compared with existing methods [8], [9], our proposed Point2Graph framework solely use the scene point cloud as input to generate open-vocabulary 3D scene graph.

## I. INTRODUCTION

3D scene graphs provide a structured scene representation by organizing rooms, objects, and their relationships in a spatial hierarchy [1]. This helps in understanding not just the presence of rooms and objects but also their locations, orientations, sizes, and relative to one another [2]. Understanding the room and object information in 3D indoor environments is essential in various domains, such as robot mapping, navigation, and path planning [3], [4]. The recent success of Large Language Model (LLM) and Visual Language Model (VLM) demonstrates that open-vocabulary capabilities are crucial for 3D scene graphs. This feature effectively connects robots and humans and facilitates robotic autonomy in complex indoor environments [5]–[7].

One of the main challenges for both closed and open vocabulary 3D scene graph generation is the scarcity of large-scale 3D-text datasets, especially compared to the abundance of large internet-scale 2D-text datasets [10]. This scarcity is even exacerbated by the fact that many available 3D-text datasets are limited to simulation environments [11]. Lacking 3D-text pair data leads to poor performance in 3D object and room segmentation and classification. Therefore, shown in Fig. 1a, most of the current literature in generating open vocabulary 3D scene graphs such as HOV-SG [8] and ConceptGraphs [9] relies on well-aligned RGB-D images and 3D point clouds to perform feature distillation or projecting features of RGB-D images to 3D point cloud to create 3D-text pairs. However, these algorithms have a requirement that well-aligned posed RGB-D images are available, which is usually challenging to achieve in real-life scenarios either because the RGB-D images are missing or the camera pose information is unavailable [12]. For example, the point clouds created from a Building Information Model (BIM) or LiDAR sensors often lack the RGB-D images and their pose data [13], [14]. Moreover, detecting and aligning 2D RGB-D images with a 3D point cloud requires the RGB-D images to have a clear view of all objects from the correct angles to accurately detect the 3D objects. However, this is often hard to achieve in real-life scenarios due to frequent issues with object occlusion and varying orientations [15]–[17].

To eliminate the hard-achieving requirement of the current open-vocabulary 3D scene graph generation, as shown in Fig. 1b, we propose *Point2Graph*. This framework only utilizes 3D point clouds as input to generate an open-vocabulary 3D scene graph. Generally speaking, our framework consists of a room segmentation and classification module and an object detection and classification module. At the room level, we enhance segmentation performance by combining a geometry-based border detection algorithm with transformer-based region detection and forming a "Snap-

Funding for V. Kamat, C. Menassa and Y. Xu was provided by NSF Award No. 2124857.

The authors are with the University of Michigan, Ann Arbor, MI 48109, USA. {yfx, luozm, qweiw, vkamat, menassa}@umich.edu

*Equal contribution

Lookup" framework for open-vocabulary room classification. At the object level, we develop an end-to-end 3D object detection and classification pipeline to detect and classify 3D objects accurately.

In particular, this work has the following contributions:

1) We propose an end-to-end point cloud-based open vocabulary 3D scene graph pipeline without the need for well-aligned posed images.
2) We present a novel room segmentation and classification pipeline that possesses the SOTA ability to comprehend highly complex and real-world scenes.
3) Our proposed point cloud-based 3D object detection and classification pipeline can achieve better performance compared with previous SOTA algorithms.
4) 4)The performance is evaluated across a range of widely used scene benchmarks and tested in navigation performance on an assistive mobile robot in real-world settings.

## II. RELATED WORK

### A. 3D Scene Graphs

3D scene graphs have been proven to be a space and time-efficient way to represent large-scale indoor scenes [2], [4], [18] by representing all rooms and objects as nodes and their belonging relationship as edges. All nodes both have their geometry (e.g., mesh, point cloud) and semantic (e.g., labels) attributes, which usually can be inferred during the robot navigation and planning process [19]. Early work such as Hydra [20] and Kimera [21] has shown how to hierarchically build a 3D scene graph containing room and object information within close-vocabulary settings. However, since these works generate 3D scene graphs in a pre-defined set of rooms and objects, their performance can be limited and impacted when they meet unfamiliar object types or adapt to human's natural instructions.

Most recently, there emerging some work such as ConceptGraph [9] and HOV-SG [8] generating 3D scene graphs in open-vocabulary settings. However, due to the lack of a 3D-text pair dataset, most open-vocabulary segmentation and classification algorithms are limited to 2D images such as SAM [22], [23]. To bridge this gap, these works project the features extracted from well-aligned RGB-D images to 3D point clouds to create 3D scene graphs. In real-life scenarios, the well-aligned posed RGB-D images are usually hard to access [12]. To overcome this limitation, we propose *Point2Graph*, which only takes the 3D scene point cloud as a single input to construct a 3D open-vocabulary scene graph in indoor environment settings.

### B. Open-vocabulary Room Segmentation and Classification

Current room segmentation methods generally fall into two categories: geometry-based and learning-based. Geometry-based methods, such as those in [8], [20], [24]–[26], identify wall borders using geometric operations on density maps. While these methods are adaptable to various scenes, their accuracy heavily relies on precise wall border extraction, which can be challenging in diverse scenarios. Additionally, other researchers leverage learning-based methods, such as HEAT [27] and Room Former [28], to perform region detection after extensive training. However, these methods lack parameter flexibility and depend on accurate border features as well. Their performance tends to be relatively poor in complex scenes with unclear borders. To improve this, our approach integrates a border-enhanced density map into the room classification pipeline, aiming to maintain region completeness while minimizing noise.

In addition, for room classification, most current room inference algorithms rely on using LLM to infer room types based on object types belonging to each room [9], [29]. However, relying solely on text-based object type information may not lead to accurate predictions if the objects are few or commonly found (such as a room containing only a chair or a table). To address this issue, we propose the "Snap-Lookup" classification pipeline, which improves room classification accuracy by extracting visual features of the rooms. These features provide more detailed information than relying solely on object types.

### C. Open-vocabulary Object Detection and Classification

Recently, 3D open vocabulary understanding has made significant progress in object-level understanding. Many methods like Openins3D [12], Uni3D [30], and ReCon++ [31] have been proposed to classify the point cloud of objects at the semantic and instance level. However, these methods commonly underperform scene-level understanding since objects in the scene are often overlapped, occluded, or incomplete. For scene-level understanding, newly proposed methods such as OpenMask3D [32], ConceptGraphs [9], and Open3DIS [33] have demonstrated the potential of using both well-aligned RGB-D images and 3D point clouds to achieve this. However, these methods still rely heavily on the availability of well-aligned 2D and 3D data pairs, which can be a limiting factor in real-world applications where such data is often incomplete or unavailable. To address these challenges, our *Point2Graph* framework proposes a unified end-to-end pipeline that begins with room-level segmentation and progresses through object detection and open-vocabulary classification, all directly from 3D point clouds. By eliminating the dependency on RGB-D image alignment, our approach enhances the robustness and flexibility of scene-level understanding in complex environments.

## III. METHODOLOGY

In this section, we present our design of *Point2Graph*, which builds a compact and enriched open-vocabulary 3D scene graph with solely 3D scene model input. The overall pipeline of *Point2Graph* is shown in Fig. 2. Generally speaking, our *Point2Graph* pipeline contains Room Segmentation and Classification (Sec. III-A) to get the room information in the scene and Object Detection and Classification (Sec. III-B) to get object information in each room. Additionally, we create a Voronoi-based navigation graph (Sec. III-C) for the robot to navigate across a large-scale environment.

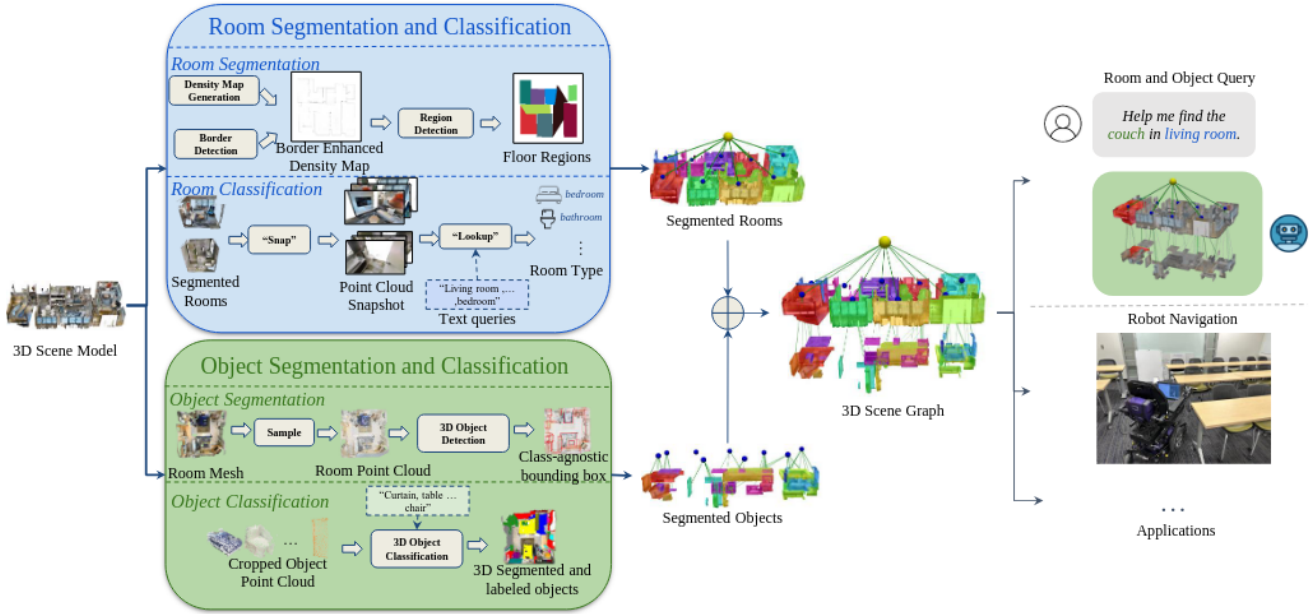

Fig. 2: **The overall pipeline of *Point2Graph***: The system is divided into two levels: the room level and the object level. At the room level, the original 3D model of the scene is used to generate a border-enhanced density map. A region detection module is then applied to segment the scene into individual rooms. Once the 3D model of each room is obtained, the "Snap-Lookup" module is used to determine the type of each room. At the object level, a sampling module is employed to extract point cloud samples for each room, and a 3D object detector is used to generate class-agnostic bounding boxes for the objects. These objects are then processed with a text query in an open-vocabulary 3D object classification module to obtain segmented and labeled objects. Finally, the information from both the room and object levels is combined to create the 3D scene graph.

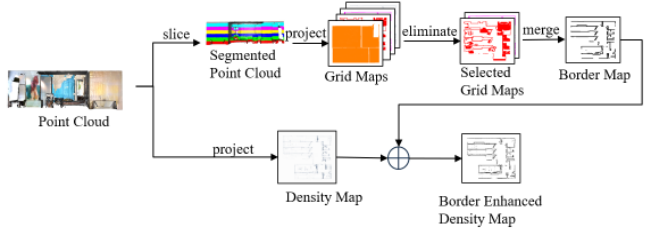

Fig. 3: **Generation of border-enhanced density map**: The process begins by segmenting the point cloud into $N$ layers, each of which is projected into a grid map. Layers that do not meet specific criteria are eliminated. The remaining valid layers are then merged to form the border, which is combined with the original density map to create a border-enhanced density map.

### A. Room-Level Segmentation and Classification

In order to get the geometry and semantic room information, our pipeline is divided into two parts: Room-Level Segmentation and Classification, which are discussed below:

*1) Room-Level Segmentation:* In order to segment each room accurately, as shown in Fig. 3, we apply a geometry-based method to generate a border-enhanced density map. First, the input point cloud is segmented into $N$ slices along the z-axis, with each slice projected onto an occupancy grid map denoted as $G_k, k = 1, ..., N$. We calculate the free space area of each grid map, denoted as $S_k, k = 1, ..., N$. To reduce noise, we eliminate grid maps that lack sufficient border information or contain excessive noise. The detailed coverage criteria are outlined in (1).

$$\mathbf{G}_{\text{select}} = \{G_k \mid \delta_b S < S_k < \delta_t S, \ k = 1, ..., N\} \quad (1)$$

where $\mathbf{G}_{\text{select}}$ is the selected set of projected grid maps. The parameters $\delta_b$ and $\delta_t$ are empirically chosen as $\frac{1}{15}$ and $\frac{1}{5}$, respectively. These selected grid map layers are then merged to construct the border map denoted as $G_{border}$, shown in (2):

$$G_{border}(i,j) = \begin{cases} 1 & \text{if } \sum_{k=1}^{M} G_k(i,j) \geq \frac{3}{4}M \\ 0 & \text{otherwise} \end{cases}. \quad (2)$$

where $M$ represents the number of valid grid maps after elimination. This step ensures that only the regions consistently identified across most layers are considered as wall boundaries.

To generate the border-enhanced density map $G_{combine}$, the border map is combined with the density map $G_{den}$ projected directly from original point cloud:

$$G_{combine} = \gamma G_{den} + (1-\gamma) G_{border} \quad (3)$$

where $\gamma$ is a weight parameter that is empirically chosen as 0.9. This process emphasizes the wall boundaries, facilitating accurate region detection.

For region detection, we employed the latest SOTA method, RoomFormer [28], as the region detector. Building on the original training provided by the paper on 3,000 scenes from Structure3D [34], we further fine-tuned the model with 100 scenes from Matterport3D (MP3D) dataset [35].

*2) Room-Level Classification:* After we got the segmented rooms, inspired by [12], we applied a "Snap-Lookup" pipeline to get the room label for each room in open-vocabulary settings. To get the 2D image for VLM to identify the types, we applied a "Snap" module as shown in Fig. 4a.

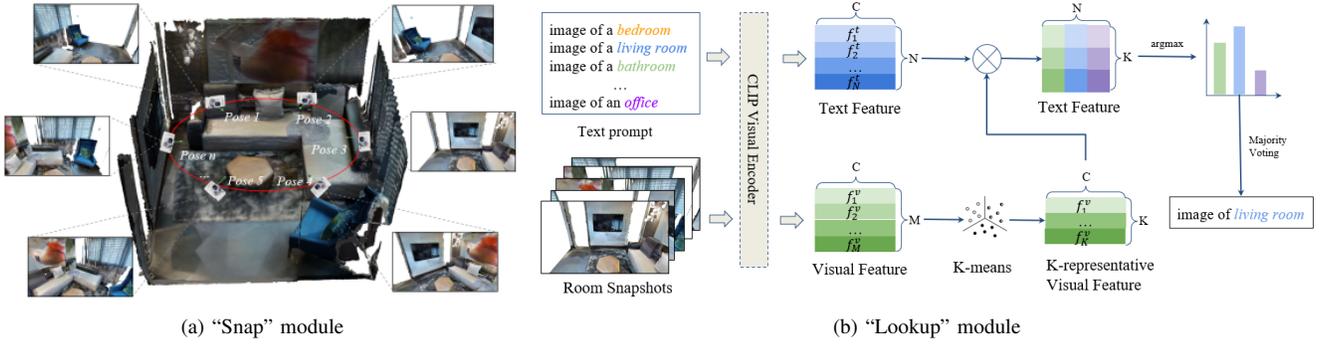

(a) "Snap" module      (b) "Lookup" module

Fig. 4: **"Snap" and "Lookup" module in room open-vocabulary classification**: (a) In "Snap" module, cameras are positioned evenly at the ellipse shape trajectory facing to the center of the room and snapshot the room from different positions. (b) In the "Lookup" module, we associate each room with open-vocabulary embeddings extracted from K-representative CLIP features. Then, argmax and majority voting are applied to get the type for each room.

We designed the camera pose $p_{camera} = (x, y, z)$ shown in (4):

$$\frac{4(x - x_c)}{L^2} + \frac{4(y - y_c)}{W^2} = 1 \qquad (4)$$
$$z = z_c$$

where the room's length and width is $(L, W)$ and its center at $p_{center} = (x_c, y_c)$. We evenly position the camera in an elliptical trajectory around the room and let the camera face the center to take snapshots from different positions.

To obtain open-vocabulary features for each room, inspired by the approach in [8], we use the CLIP visual encoder to extract embeddings from the images. Since some images may be taken from unfavorable angles, potentially confusing the final room type classification, we distill these features by extracting $K$ representative view embeddings using the K-means algorithm. Next, we construct a cosine similarity matrix between the $K$ representative features and the text features from CLIP and take the argmax along the category axis, resulting in $K$ room type predictions. Finally, we use majority voting among these predictions to determine the room's type. The "Lookup" module is illustrated in Fig. 4b.

### B. Object-Level Detection and Classification

After getting the segmentation result for each room, our approach mainly consists of two steps, where the first step deals with object detection and localization and the second step with classification. Our pipeline overview is shown in Fig. 5.

*1) Object-Level Detection:* The room point cloud data is sampled from the segmented room-level geometry information of the scene. To identify object candidates, we leverage a transformer-based module from a pre-trained 3D object detection model, V-DETR [36] to generate $N$ bounding box for each object candidate. For each proposed 3D bounding box, we extract and crop the corresponding point cloud for each object from the original room point cloud denoted as $P_b^k, k = 1, ..., N$. Since the bounding boxes may not always tightly fit the object, leading to the inclusion of irrelevant background points, we employ DBSCAN [37] to filter the cropped point cloud in order to exclude noisy points or background regions shown in (5).

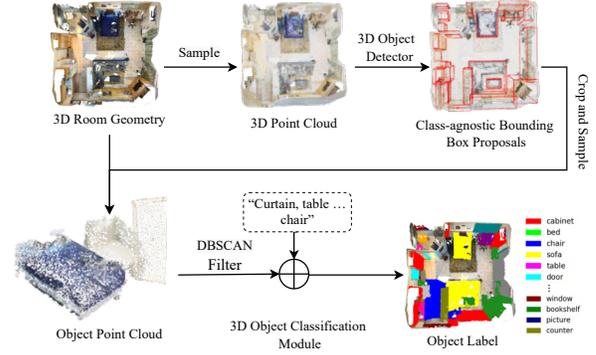

Fig. 5: **Overview of the 3D open-vocabulary detection pipeline**: It consists of two stages: (1) detection and localization using class-agnostic bounding boxes and DBSCAN filtering for object refinement, and (2) classification via cross-modal retrieval, connecting 3D point cloud data with textual descriptions, without requiring annotations or RGB-D alignment.

$$P_{obj}^k = DBSCAN(P_b^k, n), k = 1, ...N \qquad (5)$$

where $n$ represent the number of points needed for 3D object detection module and $P_{obj}^k$ represent the filtered object point cloud which contains $n$ points. After filtering, the remaining points are centered and normalized, which ensures consistency across the point cloud samples and prepares the data for the subsequent classification stage.

*2) Object-Level Classification:* In the classification stage, we perform open-vocabulary 3D object classification by leveraging a SOTA language-aligned large-scale 3D foundation model, Uni3D [30]. Specifically, the model takes as input both the filtered 3D point cloud and a textual description and retrieves the appropriate object label by identifying the alignment between the visual feature of the point cloud and the text feature. By eliminating the need for annotated training data and RGB-D alignment, our system generalizes effectively across various objects and environments without task-specific fine-tuning or manual intervention.

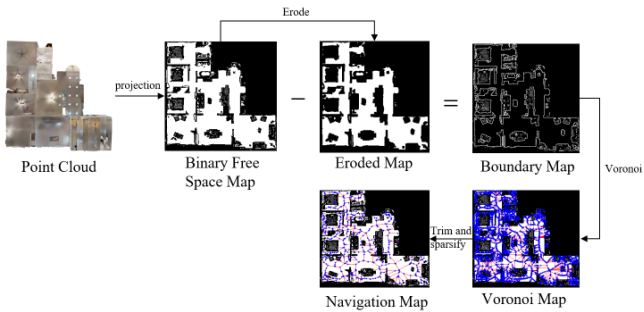

Fig. 6: **Generation of the Voronoi navigation graph**: By deducting the projected binary free space map with the eroded map, we can obtain the boundary of the environment. Then, we can construct Voronoi navigation graph and sparsify it to get our proposed navigation graph.

*C. Voronoi Navigation Graph*

In order to let a robot navigate in the area where our scene graph is built, we propose a navigation graph based on Voronoi graph [38]. As shown in Fig. 6, to get the binary free space map, we project the point cloud in $xy$-plane within the height range of $[y_{min}, y_{min} + h_{robot}]$ where $y_{min}$ is the minimal height of the point cloud and $h_{robot}$ is the height of the robot we use in our experiment. By subtracting the original binary free space map from the eroded free space map, we can get the boundary of the whole floor. Using the Voronoi planner, we are able to generate the Voronoi navigation map. We sparsify and trim it to get our final navigation graph.

## IV. EXPERIMENTAL RESULTS

*A. Evaluation for Room Segmentation and Classification*

*1) Room Segmentation Evaluation:* We conducted our experiment on the MP3D dataset [35], where all multi-floor scenes were segmented into single-floor scenes, resulting in a total of 189 scenes. Excluding those used for fine-tuning the region detector, we selected 43 scenes to serve as the test set for room segmentation. We compared our method to RoomFormer [28], the current SOTA in learning-based algorithms, and the room segmentation techniques employed in HOV-SG [8], the SOTA in geometry-based algorithms. Our evaluation metrics included Average Precision at 50% overlap (AP50) and mean Intersection over Union (mIoU).

In our experimental results, shown in TABLE I, by generating a border-enhanced density map before input to RoomFormer, our approach achieved 12% improvements in AP50 and 3% in mIoU. This demonstrates the potential of our method as an effective pre-processing module for room segmentation. The qualitative comparison of room segmentation is shown in Fig. 7, showcasing the superior performance of the room segmentation pipeline.

*2) Room Classification Evaluation:* For room classification, we evaluate our "Snap-Lookup" pipeline on 100 segmented room scenes from the MP3D dataset [35]. We classify the rooms using the total set of room categories from these scenes. As baselines, we use two methods: the zero-shot LLM-based room type inference method [29] and the room classification approach from HOV-SG [8], a leading open-vocabulary scene graph generation method. The baselines are split into two types: *privileged*, which uses ground truth object types for LLM inference, and *unprivileged*, which relies on object types detected via our proposed object detection algorithm (Section III-B). We use GPT-3.5-turbo [39] and GPT-4 [40] as the LLM baselines. Evaluation metrics include Precision, Recall, weighted F1 score, and mean Average Precision (mAP).

| Methods | AP50 | mIoU |
|---|---|---|
| HOV-SG [8] | 0.06 | 0.23 |
| RoomFormer [28] | 0.41 | 0.56 |
| **Ours** | **0.53** | **0.59** |

TABLE I: **Comparison of different room segmentation methods on the MP3D dataset.** "AP50" assesses the Average Precision calculated when the IoU threshold is set to 0.50. "mIoU" is the average of the IoU values across all segmented rooms.

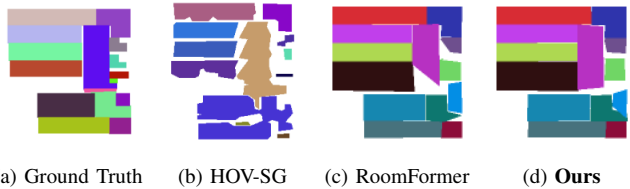

(a) Ground Truth  (b) HOV-SG  (c) RoomFormer  (d) **Ours**

Fig. 7: **Qualitative result of room segmentation**: Room segmentation produced by RoomFormer [28], HOV-SG [8] and our proposed methods on MP3D dataset [35]. Each color block represents a segmented room.

Our proposed "Snap-Lookup" pipeline, which incorporates room visual features into type inference, can differentiate between various types of rooms that contain the same objects—something text-only inference methods like GPT-3.5-turbo and GPT-4 cannot achieve. Additionally, because our "Snap" module captures snapshots of the entire room, it can obtain higher-quality images and is not restricted to specific positions and angles within the room's free space, as was the case with HOV-SG. As shown in TABLE II, our proposed "Snap-Lookup" framework yields a significant improvement among all *privileged* and *unprivileged* baselines.

*B. Evaluation for Object Detection and Classification*

We conducted our experiments on the widely-used ScanNetv2 [45] indoor point cloud dataset, which consists of 312 validation scenes, each annotated with semantic and instance segmentation masks across 18 object classes. Our pipeline is tested on the validation set for open-vocabulary instance segmentation and use the categories setting of PLA [41], excluding the "other furniture" class in ScanNetv2. We compared our method against PLA [41] and its follow-up works, RegionPLC [42] and Lowis3D [43]. To ensure fairness, we followed their category splits and reported results on novel classes. Additionally, we evaluated our method in a zero-shot inference setting against OpenIns3D [12]. We report AP at different IoU thresholds, specifically AP25 (0.25 IoU) and AP50 (0.5 IoU).

The quantitative results are summarized in TABLE III. Specifically, our method achieves the highest AP50 and

|  | Methods | Precision | Recall | F1 | mAP |
|---|---|---|---|---|---|
| Privileged | GPT-3.5-turbo w\ GT object | 0.69 | 0.61 | 0.61 | 0.63 |
|  | GPT-4o w\ GT object | 0.68 | 0.63 | 0.61 | 0.63 |
| Unprivileged | GPT-3.5-turbo w\ object detected | 0.10 | 0.14 | 0.10 | 0.14 |
|  | GPT-4o w\ object detected | 0.18 | 0.19 | 0.15 | 0.19 |
|  | HOV-SG [8] | 0.51 | 0.20 | 0.22 | 0.19 |
|  | **Ours** | **0.69** | **0.70** | **0.66** | **0.69** |

TABLE II: **Comparison of room classification method on MP3D dataset.** "Precision" measures the accuracy of positive predictions. "Recall" assesses the model's ability to find all relevant cases. The "F1" score takes their harmonic mean of Precision and Recall. "mAP" measures the average accuracy of classification across all categories

| Method | B/N | AP50 | AP25 | use RGB-D |
|---|---|---|---|---|
| PLA [41] | 10/7 | 0.22 | - | ✓ |
| RegionPLC [42] | 10/7 | 0.32 | - | ✓ |
| Lowis3D [43] | 10/7 | 0.31 | - | ✓ |
| Mask3D-P-CLIP [44] | -/7 | 0.04. | 0.08 | × |
| OpenIns3D [12] | -/7 | 0.28 | 0.43 | × |
| **Ours** | -/7 | **0.38** | **0.44** | × |
| Mask3D-P-CLIP [44] | -/17 | 0.04 | 0.14 | × |
| OpenIns3D [12] | -/17 | 0.29 | **0.39** | × |
| **Ours** | -/17 | **0.30** | 0.34 | × |

TABLE III: **Comparison of methods on ScanNetv2 dataset**. It reports Average Precision at 0.5 IoU thresholds (AP50) and 0.25 threshold (AP25). Results are compared on base/novel (B/N) category splits, with an indication of whether methods use RGB-D images ("use RGB-D")

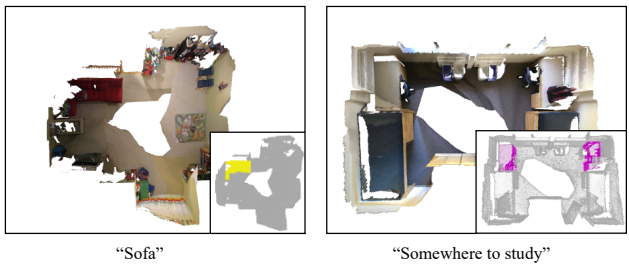

"Sofa"  "Somewhere to study"

Fig. 8: **Qualitative results of object segmentation and classification**: The results demonstrate our object segmentation and classification pipeline has ability to handle a versatile object vocabulary in diverse environments.

AP25 scores on the seven novel category predictions, as well as the top AP50 score on the seventeen novel category predictions. Our pipeline leverages the robust performance of a pre-trained 3D object detection model alongside refined object point cloud processing. Through experimentation, we observed that our object detection module accurately locates objects, while point cloud classification accuracy is enhanced by using DBSCAN to filter out noisy and background points. This integrated approach effectively captures both global and local contexts, leading to improved precision and generalization across various IoU thresholds. Additionally, we present the qualitative results in Fig. 8, showcasing object retrieval in complex 3D scenes.

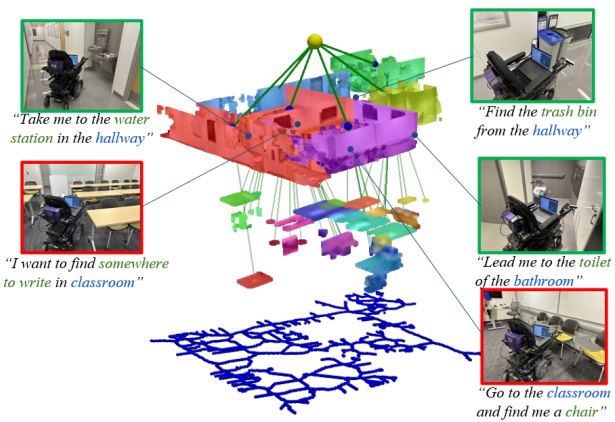

Fig. 9: **The real world experiment**: We construct the 3D open-vocabulary scene graph using *Point2Graph* algorithm and construct the Voronoi navigation graph. Then we conduct navigation tests in this environment.

### C. Real-World Experiment

To validate the system in the real world, we utilize a smart wheelchair mobile platform with a calibrated LiDAR and camera. The experiment is carried out on the first floor of the GG Brown Building. We first prompt-engineer an interpreter using GPT-4 to extract the room and object information. The robot should interpret the human's query and search for the room and object location in the pre-built 3D scene graph hierarchically and navigate to the most relevant location. For example, when users input their prompt like `Take me to the water station in the hallway`, the robot should search for `hallway` location in the room-level and `water station` location in the object-level. The qualitative result of the real-world experiment is shown in Fig. 9. We did 5 cases tests with 3 successful cases and 2 failed cases. The failed cases occur when querying the `classroom` and `somewhere to write`. Since our environment has more than one classroom and our 3D scene graph does not consider further identification for rooms with the same type, it will cause confusion for the robot.

### V. CONCLUSION

In conclusion, this work presents the *Point2Graph* framework, which addresses the limitations of current open-vocabulary 3D scene graph generation methods by eliminating the need for well-aligned posed 2D images. Through evaluations on widely used scene benchmarks and real-world tests on an assistive mobile robot, we demonstrate the robust performance in diverse and complex environments. This advancement paves the way for more adaptable and intuitive human-robot interactions in real-world scenarios.

Nevertheless, *Point2Graph* has its limitations. As discussed in the Sec. IV-C, in cases where there are multiple rooms or objects in the same environment, it will cause confusion to our open-vocabulary query in 3D scene graph. To overcome this situation, integrating deep-learning algorithm for room number detection [46] or using multi-round LLM-based conformal prediction [47] to get a clearer human's intention for query are promising future directions.